# Enhancing Bankruptcy Prediction of Banks through Advanced Machine Learning Techniques: An Innovative Approach and Analysis

Zuherman Rustam[1]*, Sri Hartini[1], Sardar M.N. Islam[2], Fevi Novkaniza[1], Fiftitah R. Aszhari[1], Muhammad Rifqi[3]
[1] Department of Mathematics, Universitas Indonesia, Depok 16424, Indonesia
[2] Institute for Sustainable Industries and Liveable Cities, Victoria University, Melbourne 14428, Australia
[3] Research Group, Indonesia Deposit Insurance Corporation, Jakarta 12190, Indonesia
*Corresponding emails: rustam@ui.ac.id; Sardar.Islam@vu.edu.au

**Abstract**

**Context:** Financial system stability is determined by the condition of the banking system. A bank failure can destroy the stability of the financial system, as banks are subject to systemic risk, affecting not only individual banks but also segments or the entire financial system. Calculating the probability of a bank going bankrupt is one way to ensure the banking system is safe and sound. **Existing literature and limitations:** Statistical models, such as Altman's Z-Score, are one of the common techniques for developing a bankruptcy prediction model. However, statistical methods rely on rigid and sometimes irrelevant assumptions, which can result in low forecast accuracy. Improved approaches are necessary. **Objective of the research:** Bankruptcy models can be developed using machine learning techniques, such as logistic regression (LR), random forest (RF), and support vector machines (SVM). These machine learning techniques are chosen because they are straightforward to implement and do not require rigid assumptions such as multivariate normality and equal variance-covariance matrices for each group. According to several studies, machine learning is also more accurate and effective than statistical methods for categorising and forecasting banking risk management. This paper aims to develop a generalised model capable of forecasting bank bankruptcies based on data from both commercial and rural banks. The likelihood associated with each model is also examined in this study. **Present Research:** The commercial bank data are derived from the annual financial statements of 44 active banks and 21 bankrupt banks in Turkey from 1994 to 2004, and the rural bank data are derived from the quarterly financial reports of 43 active and 43 bankrupt rural banks in Indonesia between 2013 and 2019. Five rural banks in Indonesia have also been selected to demonstrate the feasibility of analysing bank bankruptcy trends. **Findings and implications:** The results of the research experiments show that RF can forecast data from commercial banks with a 90% accuracy rate. Furthermore, the three machine learning methods proposed accurately predict the likelihood of rural bank bankruptcy. However, it is recommended that RF and SVM be combined for the most effective prediction results. **Contribution and Conclusion:** The proposed machine learning method is generally capable of making accurate predictions for bank bankruptcies. It can also help to implement policies that reduce the costs of bankruptcy or for authorities to act to maintain economic stability if early warnings about bank failures are provided.

## 1. Introduction

**Context:** In both Indonesia and Turkey, the banking sector dominates the country's financial system as a whole. The banking sector accounts for nearly 70% of the total financial system assets in Indonesia [1] and 91% of the total financial sector assets in Turkey [2]. This means that the stability of the entire financial system is considerably affected by the condition of the banking system. In Indonesia, there are two main types of banks: commercial banks and rural banks. Despite running similar basic banking practices such as collecting deposits and disbursing loans, commercial and rural banks differ in several characteristics. The first is the ability to provide foreign currency payment services. Commercial banks can provide foreign currency payment services (such as buying and selling of foreign currency) but rural banks do not. The second is the form of deposits collected from the public. In addition to savings and time deposits, commercial banks also collect funds in the form of demand deposits and certificates of deposit whereas rural banks can only accept savings and time deposits. The types of banks operating in Turkey are relatively limited. In addition to the most common deposit-taking commercial banks, "participation banks" operate based on Sharia principles, as do development/investment banks [3].

Rural banks play a significant role in promoting the local economy, especially among those from low-capital societies and informal business sectors [4]. Law No. 10 of 1998 states that business entities are responsible for collecting and distributing the social fund. Rural banks are able to support rural modernization and provide financial services to low economic communities / small micro-enterprises. Rural banks specifically aim to provide banking products and services for low-income households and small and micro enterprises (SMEs) in both cities and rural areas. In general, the purpose and characteristics of Sharia rural banks are relatively similar to other microfinance institutions (MFI) [5]. As the management and quality of a rural bank's operations improve, they also positively influence the growth of the real sector, especially the informal sector, which is a significant part of the economy [6].

However, a bank is exposed to the risk of failure or bankruptcy in its operation. A bank failure is defined as the closure of an insolvent bank or one that lacks the funds to pay off its debts, as determined by the regulator. Bank failures have a significant impact on the country, its citizens, and individuals. For example, the massive bank failures in Indonesia following the 1998 crisis were overwhelming and caused panic among depositors, triggering massive bank runs [7]. A crisis of comparable magnitude also happened to the Turkish banking system in 1994 and 2000 [8]. To safeguard against this possibility, a deposit guarantee system was established. There are various types of deposit guarantee systems, but all share similar purposes: to guarantee and cover as many depositors as possible, and are carried out by a credible, independent, and accountable institution. The Indonesia Deposit Insurance Corporation (IDIC) is an institution established by law to provide deposit insurance for depositors and actively participate in creating and maintaining financial system stability, in collaboration with the Minister of Finance, Bank Indonesia, and the Banking Supervisory Agency. Similarly, the Savings Deposit Insurance Fund (SDIF) / *Tasarruf Mevduatı Sigorta Fonu* (TMSF) exists in Turkey since its formation in 1983 to strengthen and restructure the financial structure of banks [8].

In Indonesia, according to Law No. 24/2004, every bank conducting business activities in the territory of the Republic of Indonesia is required to participate in the IDIC guarantee scheme to collectively establish a supervised banking system, thereby minimising the consequences of bank bankruptcy. SDIF in Turkey also requires compulsory participation from all banks operating in the Country. In addition, bankruptcy prediction can also be performed as early as possible to prevent bankruptcy and minimize the costs associated with bank bankruptcy. Furthermore, predicting the probability of bank bankruptcy is beneficial for the authorities responsible for maintaining the stability of the banking system, allowing them to take anticipatory actions.

**Existing Literature and Limitations, and the Objective and Contributions of this Paper:** One of the approaches to building a bankruptcy prediction model is using statistical models. The utilization of statistical methods to predict corporate bankruptcy dates back to Altman's [9] Z-Score. Altman used multivariate discriminant analysis to build the model, and his study made an important contribution in the field of bankruptcy research since the bankruptcy prediction research that followed was based on his framework and utilized various statistical methods such as multiple variable discriminants, linear discriminant analysis, quadratic discriminant analysis, multiple regression, and factor analysis. With recent advances in computational science, several researchers have explored machine learning approaches to predict bankruptcy, including Barboza et al. [10], Wang [11], and Qu et al. [12].

Machine learning has been widely applied in various fields and has been shown to outperform traditional statistical approaches in classifying and predicting banking risk management [13]. Several studies have demonstrated that machine learning is more accurate and effective than statistical methods [10, 11, 14]. The primary difference between machine learning and statistical methods is that in traditional statistical methods, the researcher must first determine the model's structure and then estimate the model parameters to match the observed data. However, in machine learning, the system examines specific model structures by analysing the data. Moreover, statistical methods rely on strict (and sometimes unrealistic) assumptions (such as normal distribution and no correlation between variables), which can lead to poor prediction accuracy [15].

Despite these advancements, most studies tend to focus on developing prediction models themselves, but place less emphasis on analysing the probability entailed by the models. To fill this gap, this study aims to develop a new and improved model to predict bank bankruptcy. it also analyzes the probability associated with each model. Moreover, this study not only focuses on the empirical analysis of the prediction model but also attempts to further analyse the individual conditions of the rural banks in the sample to evaluate any idiosyncratic issues.

**Present Research:** In our study, we utilise data from the annual financial statements of commercial banks in Turkey [8] for the commercial banks, and data issued by the Indonesia Financial Services Authority (OJK) in the form of Banking Publication Reports for the rural banks. The data is processed and analyzed using machine learning methods. The machine learning methods in use are approaches that commonly achieve high accuracy in classification tasks, such as logistic regression, random forest, and support vector machines (SVMs) [16, 17]. Nevertheless, this study is considered exploratory, and further research is still needed on the discussion**.**

**Structure of the Paper:** The rest of this paper is organized as follows. Section 2 gives a brief review of commercial and rural banks in Indonesia. Section 3 discusses the machine learning algorithms and methods employed in this study, as well as the data used to implement these algorithms. Section 4 presents the results of the machine learning algorithms in predicting the bankruptcy of commercial and rural banks, including a further analysis of the individual conditions of the rural banks in the sample. Finally, the conclusion of this paper is given in Section 5.

## 2. Commercial and Rural Banks

Experts have different opinions on how to interpret the word "bank." According to the Banking Companies Act of India, a bank is a financial institution that accepts money from the public for the benefit of borrowers or investors upon request or withdrawal by check, draft, or otherwise. Banks can also be regarded as financial institutions that provide financial services to their customers, such as accepting deposits or providing loans. From the existing definitions, it can be concluded that a bank is an institution or financial institution that deals with debits and credits. Banks are the bridge between savers and borrowers. Therefore, banks are important to any country and, of course, they need to be an economic driver [18].

This study discusses two types of banks, namely commercial and rural. Commercial banks carry out business activities conventionally or based on Sharia principles to provide services in payment traffic. Commercial banks can engage in the following activities: collecting funds from the public in the form of deposits, demand deposits, time deposits, certificates of deposits, savings, or other equivalent forms; providing credit; issuing letters of debt acknowledgement; buying, selling, or guaranteeing at their own risk or for the benefit of their customers; and transferring money either for their own interests or for the benefit of their customers. They can also provide finance and/or carry out other activities based on Sharia principles, as well as activities commonly carried out by banks, provided they do not conflict with Sharia law and the prevailing laws and regulations.

Rural banks carry out business activities conventionally or based on Sharia principles, which does not include services related to payment transactions. The business of a rural bank includes collecting funds from the public in the form of deposits, in the form of time deposits, savings, and/or other equivalent forms; providing credit; providing financing and placement of funds based on Sharia principles; and placing the funds in the form of time deposits, certificates of deposit, and/or savings in other banks. The business activities of rural banks are narrower than those of commercial banks because there are several restrictions on rural banks, namely: conducting insurance business; carrying out capital participation; carrying out business activities in the form of foreign exchange; accepting deposits in the form of current accounts; and participating in running payment traffic.

## 3. Data and Methodology

This section discusses the data and methods used in predicting bank bankruptcy.

### *a. Data*

This study examines two types of datasets: commercial bank data and rural bank data. The commercial bank data is the annual financial report data of *Banks in Turkey* for the period 1994-2004 issued by the Bank Association of Turkey (BAT). The data is derived from Boyacioglu et al. [8], which consists of 44 active banks (not bankrupt) and 21 bankrupt banks. The status of a bank (whether it is bankrupt or not) is analysed and predicted based on the calculation of financial ratios derived from its financial statements. For this commercial bank dataset, we utilise the CAMELS ratio, which comprises a total of 20 predictors, as shown in Table 1.

Table 1. Variables of Commercial Bank Data

| CAMELS Ratio | Variables | Description |
|---|---|---|
| *Capital adequacy* | CA1 | *Shareholder's equity/total assets* |
| | CA2 | *Shareholder's equity/total loans* |
| | CA3 | *Shareholder's equity + net profit/total assets + off balance sheet commitments* |
| *Asset quality* | AQ1 | *Permanent assets/total assets* |
| | AQ2 | *Total loans/total assets* |
| | AQ3 | *Loans under follow-up/total loans* |
| | AQ4 | *Specific provision/total loans* |
| | AQ5 | *Specific provision/total loans* |
| *Management* | M1 | *Personnel expenses/average assets* |
| *Earnings* | E1 | *Net profit/average assets* |
| | E2 | *Net profit/average shareholder's equity* |
| | E3 | *Income before taxes/average assets* |
| | E4 | *Interest income/total operating income* |
| | E5 | *Non-interest expenses/ total operating income* |
| *Liquidity* | L1 | *Liquid assets/total assets* |
| | L2 | *Total loans/total deposits* |
| *Sensitivity to market risk* | SMR1 | *Trading securities/total assets* |
| | SMR2 | *FX assets/FX liabilities* |
| | SMR3 | *Net interest income/average assets* |
| | SMR4 | *Net balance sheet position/total shareholder's equity* |

The rural bank data used in this study were obtained from the quarterly financial reports of Indonesian conventional rural banks for the 2013-2019 period, which consists of 43 bankrupt rural banks and 43 active rural banks. The data is issued by the Indonesian Financial Services Authority (OJK). For the rural bank data, we derived 5 predictor variables, also in line with the CAMEL principle, as displayed in Table 2.

Table 2. Variables of Rural Bank Data

| Variables | Description |
|---|---|
| *Capital adequacy* | *Capital Adequacy Ratio* (CAR) ratio, i.e. the capital to Risk Weighted Assets Ratio |
| *Asset quality* | Earning Assets Quality ratio, i.e. the ratio of classified earning assets to total earning assets. |
| *Management* | *Net Profit Margin* (NPM) ratio, i.e. the ratio of net income to operating income. |
| *Earnings* | *Return on Asset* (ROA) ratio, i.e. the ratio of net income to total assets. |
| *Liquidity* | *Loan Deposit Ratio* (LDR) ratio, i.e. ratio of total lending to total deposit. |

Furthermore, based on the data from 43 rural banks data which have a bankruptcy status, 4 banks were randomly selected to obtain 10 financial reports for the last quarter before the bank was liquidated. These data were used in the analysis of trends in bankruptcy probability prediction.

*b. Data Preprocessing*

The preliminary analysis revealed that there are several missing values. In this case, the data cleaning process is carried out to remove missing values. For the dependent variables, we assign a value of 0 if the bank is active and 1 if it is bankrupt.

Moreover, there is an inequality in the amount of data between the two categories for commercial banks, i.e. 44 active banks and 21 bankrupt banks. Therefore, we utilise a resampling method named the Synthetic Minority Oversampling Technique (SMOTE) to solve the problem of unbalanced commercial bank data. In practice, SMOTE balances the dataset by adding a number of samples to the minority class [19]. The steps for balancing data using SMOTE are as follows [20]:

1. Determine the k-nearest neighbours from the data sample for the minority class.
2. Choose a number between 0 and 1 randomly.
3. Calculate the difference value between the minority class sample and its nearest neighbours.
4. Multiply the result in step 3 by the number in step 2.
5. Add the result in step 4 to the minority class sample.
6. The resulting data sample is synthetic.

Furthermore, using balanced commercial bank data and rural bank data, the data for each bank is divided into a training and testing dataset, where 75 percent of the data is used for training and 25% is used for testing. The composition of the training data and testing data samples from commercial banks and rural banks is given in Table 3.

Table 3. Composition of Training Data and Testing Data from Commercial Banks and Rural Banks

| Data | Training Data | | Testing Data | |
|---|---|---|---|---|
| | Active Bank | Bankrupt bank | Active Bank | Bankrupt bank |
| Commercial bank | 33 | 33 | 11 | 11 |
| Rural bank | 30 | 34 | 13 | 9 |

The machine learning algorithms used to build the model are logistic regression, random forest and support vector machines. To ensure the highest accuracy possible, hyperparameter tuning is performed to identify the optimal hyperparameters for each model. To achieve this, the grid search method was employed to test all variations of the given hyperparameters using the model's 5-fold cross-validation. In the 5-fold cross-validation process, the data is divided into five parts. Then, in turn, one part is used as testing data and the other part is used as training data. In addition, suppose model A has hyperparameters X and Y. In the grid search method, the best combination of hyperparameters is sought that yields the highest accuracy in model A, and this combination is then used in model A to test the data.

*c. Logistic Regression*

Logistic regression is a machine learning algorithm that can be used when the data has a binary and categorical response variable, in this case, "bankrupt or active". The probability of each response variable occurring can be predicted using logistic regression. Suppose there are $n$ observations $(x_i, y_i),\ i = 1,2,\ldots,n$ where $y_i$ represents the value of the response variable and $x_i$ is the value of the independent variable from the *i*-th observation. Next, assume that the response variable has been encoded in a binary format with values of 0 or 1. A simple logistic regression model is formed as follows:

$$\text{logit}(Y) = \ln\left(\frac{\pi}{1-\pi}\right) = \beta_0+\beta_1 X \quad (1)$$

where the probability of occurrence of the event $Y = 1$ is defined as

$$\pi(x) = \Pr(Y = 1) = \frac{e^{\beta_0+\beta_1 x}}{1 + e^{\beta_0+\beta_1 x}} \quad (2)$$

where $\beta_0$ and $\beta_1$ are parameters whose values are estimated using the maximum likelihood [21].

A maximum likelihood estimator is a technique used to estimate unknown parameter values by maximizing the likelihood function. The likelihood function states the probability of the observed data as a function of the unknown parameters. The contribution to the likelihood function of each pair $(x_i, y_i), i = 1,2,\ldots,n$ is given as follows:

$$\pi(x_i)^{y_i}[1 - \pi(x_i)]^{1-y_i} \quad (3)$$

Since it is assumed that each pair of observation points are independent of each other, the likelihood function of the model can be expressed as the multiplication of the probability mass function for each observation on the data expressed as:

$$l(\beta) = \prod_{i=1}^{n} \pi(x_i)^{y_i}[1-\pi(x_i)]^{1-y_i} \quad (4)$$

To simplify the calculation process, the likelihood function is formed, which is the logarithm of equation 4. This expression is known as the log-likelihood, which is defined iEquation (5).

$$L(\beta) = \sum_{i=1}^{n}\{ y_i \ln[\pi(x_i)] + (1-y_i)\ln[1-\pi(x_i)]\} \quad (5)$$

Then, the likelihood function is differentiated against $\beta_0$ dan $\beta_1$ in such that two likelihood equations are obtained.

$$\sum_{i=1}^{n}[ y_i - \pi(x_i)] = 0 \quad (6)$$

and

$$\sum_{i=1}^{n} x_i[ y_i - \pi(x_i)] = 0 \quad (7)$$

The solution of equations (6) and (7) will provide an estimate of the value of β, which can then be used in Equation (1) and Equation (2) to estimate the probability of bank bankruptcy.

*d. Random Forest*

Random Forest (RF) was introduced by Breiman [22] in 2001. This method is an extension of the classification and regression tree (CART) by applying the bagging (bootstrap aggregating) method and random feature selection. RF consists of more than one decision tree, which makes this method more robust in classification and regression [22]. RF is used to regress with the average output of all the decision trees formed [23].

The formation of multiple trees in RF is obtained from resampling results, which return the training data, a method known as the bootstrap method. The basic idea is to draw a random sample of all data with returns where each sample is the same size of the number of observed data. Let $S = x_1, x_2, \ldots, x_n$ be a random sample of size $n$ labelled $Y(x_1), Y(x_2), \ldots, Y(x_n)$. In the bootstrap method, *B* will take an $n$ size sample of the original sample, accompanied by a return [24].

The selection of the predictor variables used as the separator is taken randomly. However, the predictor variables at the higher level for each decision tree are more important than the predictor variables at the lower level. In this case, the predictor variables provide more information in the decision-making process. The Gini index is a measure commonly used in the selection of predictor variables as in its original algorithm [22].

Suppose there is a training data set $S$ with $m$ features and $k$ classes, $S(A_1, A_2, \ldots, A_m, Y)$ Dengan $Y$ is an attribute consisting of $k$ different classes. So, the Gini value can be calculated as follows [25]:

$$\mathrm{Gini}(S) = 1 - \sum_{i=1}^{k} p(y_i)^2 \qquad (9)$$

where $p(y_i)$ represents the probability (relative frequency) of $y_i$ class on $S$ training data.

The classification process on RF is run after all the bootstrap tree results are formed. The prediction of the class of each data point is carried out based on the majority vote from the prediction results of several decision trees that have been formed. Meanwhile, the average calculation is carried out to obtain class probability predictions for each data point [26]. In other words, the probability that a data point belongs to class $Y = y$ can be written as

$$\Pr(Y = y) = \frac{\text{The number of trees that produce decisions } Y = y}{\text{The number of trees formed in the RF method}} \qquad (10)$$

The RF algorithm for classification [27] and regression is:

1. Given training data $S$ with $n$ observations and $m$ predictor variables.
2. For $b = 1, 2, \ldots, B$:
    a. Perform the bootstrap stage, namely resampling $n$ size from $n$ observations on the training data with returns.
    b. Build $h^b$ A decision tree is formed from each set of bootstraps, with each node of the tree randomly selected.
        i. Select $p$ predictor variable randomly from $m$ predictor variables, with $p \leq m$.
        ii. Select the best predictor variable from $p$ based on the results of the calculation on the Gini index measure, that is $\hat{p} = A_i \mid \mathrm{Gini}(S^b, A_i) =$ the minimum value of $\mathrm{Gini}(S^b, A_l)$ for $i = 1, 2, \ldots, m$ where,
        $$\mathrm{Gini}(S^b, A_i) = \sum_{v \in A_i} \frac{|S_v^b|}{|S^b|} \mathrm{Gini}(S_v^b).$$
        iii. Perform node splitting.
    c. Build each decision tree without pruning.
3. Output of ensemble decision trees $\{h^b\}_1^B$.
4. Prediction of data point classes based on majority vote.
5. Predict the probability of data point classes using Equation (10).

*a. Support Vector Machines*

A support vector machines (SVM), introduced by Cortes and Vapnik [28] in the late 1990s, is a machine learning algorithm used for classification and regression. SVM utilises the margin concept, defined as the shortest distance between the decision boundary and any training data point. A decision boundary is a linear model or hyperplane $y(x)$ with parameters $w$ and $b$. By maximizing the margin, a certain decision boundary is obtained where the maximum margin provides the best generalization capability.

SVM uses the maximum margin separator, commonly called a hyperplane. The hyperplane is located at a maximum distance from different classes. Therefore, an optimisation step is necessary to ensure the hyperplane generalises better, as it is located at the maximum distance from each class. This is very important because, at the training step, SVM only uses a population sample, whereas when predicting, SVM uses testing data that has not been studied and may have a slightly different distribution from the trained data.

Suppose that $T = \{(x_i, y_i)\}$ is a linearly separated training dataset. $T = \{(x_i, y_i)\};\ i = 1, \ldots, N$; $x_i \in R^n$ as the dataset, $y_i \in \{-1, +1\}$ as the data class, and $N$ is the amount of data. The hyperplane formula with class -1 and +1 is shown in Equation (11).

$$w \cdot x + b = 0 \tag{11}$$

Variable $w$ represents the normal plane to the hyperplane and $b$ represents the bias. The formula in Equation (11) separates the positive and negative training data, which are formulated in Equations (12) and (13).

$$w \cdot x_+ + b \geq +1 \tag{12}$$

$$w \cdot x_- + b \leq -1 \tag{13}$$

Variable $x_+$ is a positive sample and $x_-$ is a negative sample. In binary classification, the decision function in SVM is formulated in Equation (14).

$$\mathrm{g(x) = sign(w \cdot x + b)} \tag{14}$$

Data are classified as a negative class if they are located below the decision line, namely $g(x) < 0$ and data are classified as a positive class if they are located above the decision line, namely $g(x) \geq 0$. A general illustration of SVM is provided in Figure 1, which shows the optimal hyperplane for separating data with two classes that maximises the margins.

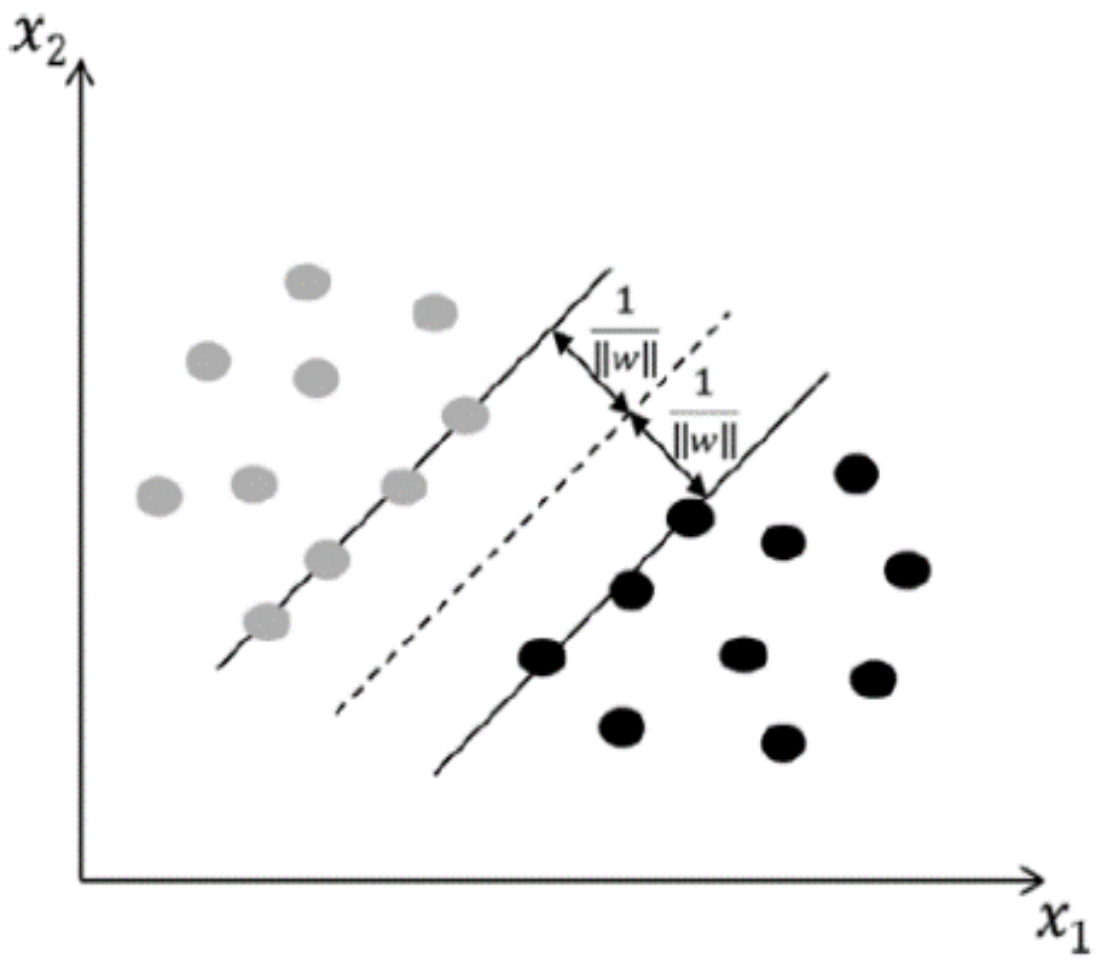

Figure 1. Illustration of SVM.

*b. Performance Measure*

This section evaluates the machine learning model's performance using an accuracy measure as a performance indicator. The prediction model yields four possible results based on the predictions obtained. The first possibility is a true positive (TP), which is a condition where the bankrupt bank was predicted correctly as bankrupt. The second possibility is a false positive (FP), a condition where the active banks incorrectly predict a bank as bankrupt. A third possibility is a true negative (TN), a condition that the active banks correctly predicted as inactive. The last possibility is a false negative (FN), a condition where the bankrupt bank incorrectly predicted itself as active. The four possible results of these predictions are summarized in the confusion matrix in Table 4.

Table 4. Confusion Matrix

| Class | | Prediction Class | |
|---|---|---|---|
| | | Bankrupt | Active |
| Actual Class | Bankrupt | TP | FN |
| | Active | FP | TN |

Furthermore, after obtaining the confusion matrix, the accuracy value of each model is calculated. Accuracy measures how well a model predicts all the data samples it is intended to predict. Using bankrupt and active bank data, accuracy provides an answer to the question of how many data samples were correctly predicted. Accuracy is calculated using Equation (15).

$$\text{Accuracy} = \frac{\text{TP} + \text{TN}}{\text{TP} + \text{FN} + \text{FP} + \text{TN}} \times 100\% \tag{15}$$

## 6. Result and Discussion

This section discusses the results of random forest, logistic regression, and support vector machines in predicting the bankruptcy of commercial and rural banks' bankruptcy.

### *a. Bankruptcy Prediction of Commercial Banks*

The random forest model achieves an accuracy of 83.52% on the training data and 90.91% on the testing data. Table 5 shows the confusion matrix derived from the testing data. According to the table, two samples of active banks are incorrectly forecasted to go bankrupt. On the other hand, all bankruptcy banks are correctly forecasted.

Table 5. Confusion Matrix of Random Forest Model in Predicting Commercial Bank Bankruptcy

| | | Predicted Class | |
|---|---|---|---|
| | | Bankrupt | Active |
| Actual Class | Bankrupt | 11 | 0 |
| | Active | 2 | 9 |

The logistics regression model's prediction accuracy is 75.82% on the training data and 77.27% on the testing data. This model has a lower accuracy than the random forest model. Table 6 illustrates the confusion matrix for the testing data. According to Table 6, two samples of active banks are incorrectly predicted to be bankrupt, a finding consistent with the results from random forests. In addition, three samples of bankrupt banks are incorrectly predicted as active banks, which means this model is less accurate than the random forest model.

Table 6. Confusion Matrix of Logistic Regression Model in Predicting Commercial Bank Bankruptcy

| | | Predicted Class | |
|---|---|---|---|
| | | Bankrupt | Active |
| Actual Class | Bankrupt | 8 | 3 |
| | Active | 2 | 9 |

The SVM model's prediction accuracy is 77.36% for the training data and 81.82% for the testing data. This model has lower accuracy than the random forest model but is more accurate than the logistic regression model. Table 7 illustrates the confusion matrix for the testing data. According to Table 7, three samples of active banks are incorrectly predicted to go bankrupt. Furthermore, one of the bankrupt bank samples is misclassified as an active bank.

Table 7. Confusion Matrix of Support Vector Machines Model in Predicting Commercial Bank Bankruptcy

| | | Predicted Class | |
|---|---|---|---|
| | | Bankrupt | Active |
| Actual Class | Bankrupt | 10 | 1 |
| | Active | 3 | 8 |

Thus, when compared to the other two proposed prediction models, the random forest model performs the best in predicting whether commercial banks are bankrupt or active. This study's research on commercial banks focuses on determining whether a bank is active or bankrupt. However, when time-series data on bank failure are available, the random forest model may be extended to calculate the likelihood of bank bankruptcy.

*b. Bankruptcy Prediction of Rural Bank*

The accuracy of the prediction results using random forest models and SVMs is 98.33% for the training data and 100% for the testing data. Meanwhile, the accuracy of the logistic regression model is 92.05 per cent for the training data and 100 per cent for the testing data. The confusion matrix for the testing data, using random forest, logistic regression, and SVM models, is shown in Table 8.

Table 8. Confusion Matrix of All Three Proposed Machine Learning Models in Predicting Rural Bank Bankruptcy

| | | Predicted Class | |
|---|---|---|---|
| | | Bankrupt | Active |
| Actual Class | Bankrupt | 9 | 0 |
| | Active | 0 | 13 |

*c. Comparison of Machine Learning Model Accuracy in Commercial and Rural Bank Data*

Table 9 compares the accuracy of random forest, logistic regression, and support vector machines. As shown in Table 9, the random forest approach outperforms the other two methods. The increased accuracy demonstrates this for both the training and testing data. Random Forest has an accuracy of 83.52% when used with training data and 90.92% when used with testing data.

Table 9. Comparison of the Proposed Machine Learning Model Accuracy in Predicting Commercial Bank Bankruptcy

| Methods | Accuracy (%) | |
|---|---|---|
| | Training data | Testing data |
| Random forest | 83.52 | 90.91 |
| Logistic regression | 75.82 | 77.27 |
| Support vector machines | 77.36 | 81.82 |

In experiments using rural bank data, in contrast to commercial bank data, the random forest model has the same accuracy for the training and testing data as the support vector machines, as illustrated in Table 10. Additionally, while logistic regression achieves 100 per cent accuracy on the testing data, it yields only 92.05 per cent accuracy on the training data. The accuracy of this model is lower than that of the other two proposed machine learning models. Thus, because random forest and support vector machine models outperform logistic regression in terms of training and testing data accuracy, these two models were chosen for further analysis of the probability of rural bank bankruptcy.

Table 10. Comparison of Proposed Machine Learning Model Accuracy in Predicting Rural Bank Bankruptcy

| Methods | Accuracy (%) | |
|---|---|---|
| | Training data | Testing data |
| Random forest | 98.33 | 100 |
| Logistic regression | 92.05 | 100 |
| Support vector machines | 98.33 | 100 |

*d. Trend Analysis in Predicting the Probability of Indonesia's Rural Bank Bankruptcy*

In Indonesia, the financial services authority can determine one of three supervisory statuses: normal supervision, intensive supervision, or special surveillance. If a rural bank is not classified as "under intensive supervision" or "under special surveillance," it is said to be under normal supervision. This supervisory system is designed to help banks avoid bankruptcy. To this day, rural banks' supervisory status is determined by two factors: (1) the minimum capital adequacy ratio; and (2) the average value of the cash ratio for the preceding six months or two quarterly financial reports prior to the determination of intensive supervision or special supervision. This study, on the other hand, employs machine learning to assess the likelihood of a rural bank's bankruptcy by examining the CAMEL ratio.

This section analyzes four samples of rural banks that have been liquidated or are in the process of being liquidated to determine the trend probability of rural bank bankruptcy. The analysis is based on the bank's quarterly financial reports for the two years preceding its license loss.

In the first sample, rural bank A is located in South Tangerang City, Banten Province. The bank's business license was revoked on November 22, 2018, after being designated as "under special surveillance" on August 29, 2018, due to a capital adequacy ratio of less than 0%. Rural bank A's trend prediction probability is depicted in Figure 2.

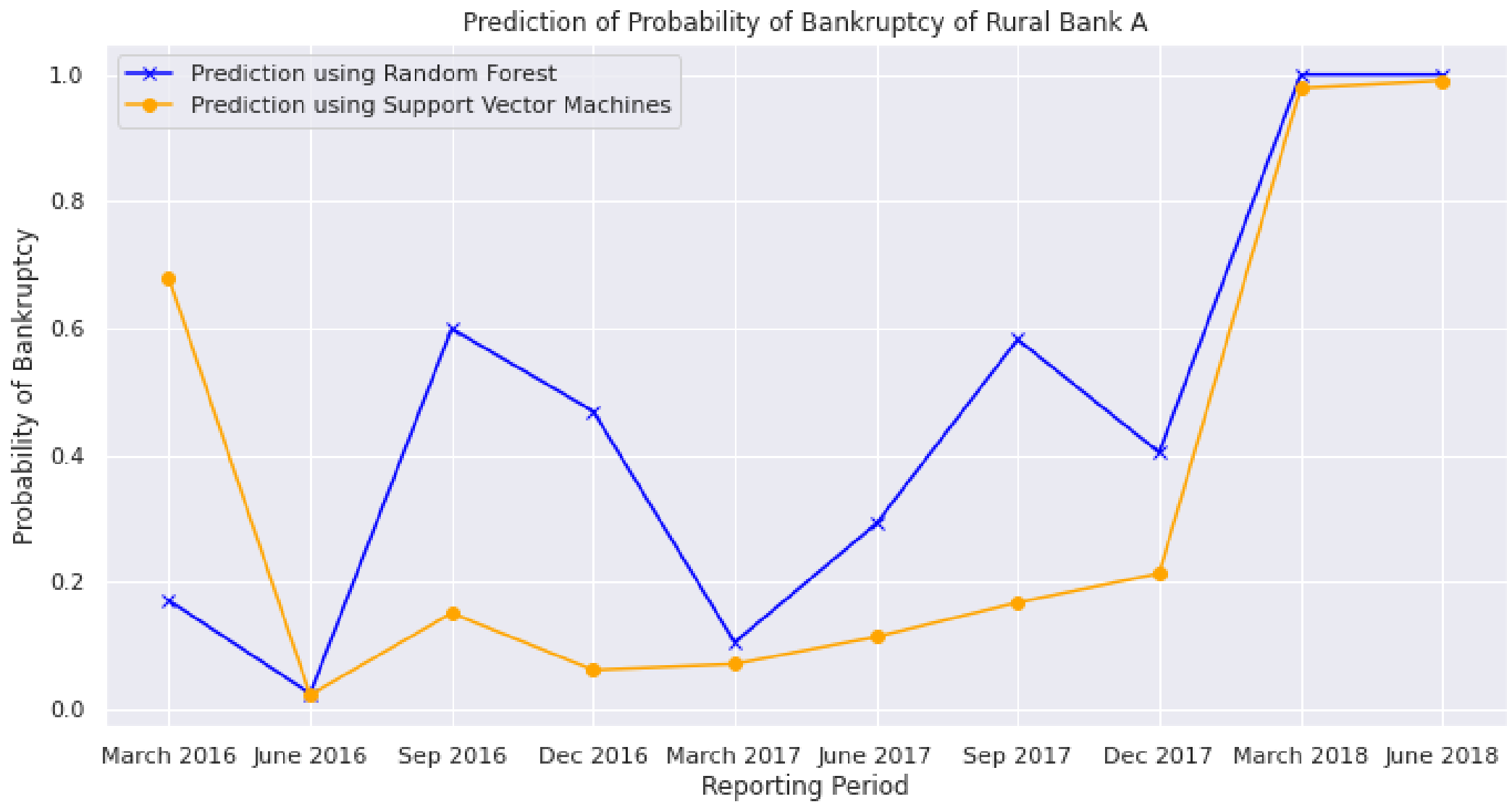

Figure 2. The Trend of Rural Bank A's Bankruptcy Probability.

According to Figure 2, this rural bank was suspected of bankruptcy in March 2016 (a probability of bankruptcy greater than 0.5), though this probability decreased significantly in the subsequent quarterly report. Then, in September 2016 and September 2017, the random forest model predicted an increase in the probability of bankruptcy. Following this, the random forest and support vector machine models forecasted bankruptcy in March 2018, five months before rural bank A was put "under special surveillance."

The second example is rural bank B, located in Bali's Badung Regency. As of August 13, 2019, the bank's business license was revoked. From May 16, 2018 to May 16, 2019, the rural bank was previously assigned the status of "under intensive supervision." On March 29, 2019, the status was changed to "under special surveillance" (which indicates a deteriorating condition) until June 29, 2019. The trend of rural bank B's prediction probability is depicted in Figure 3.

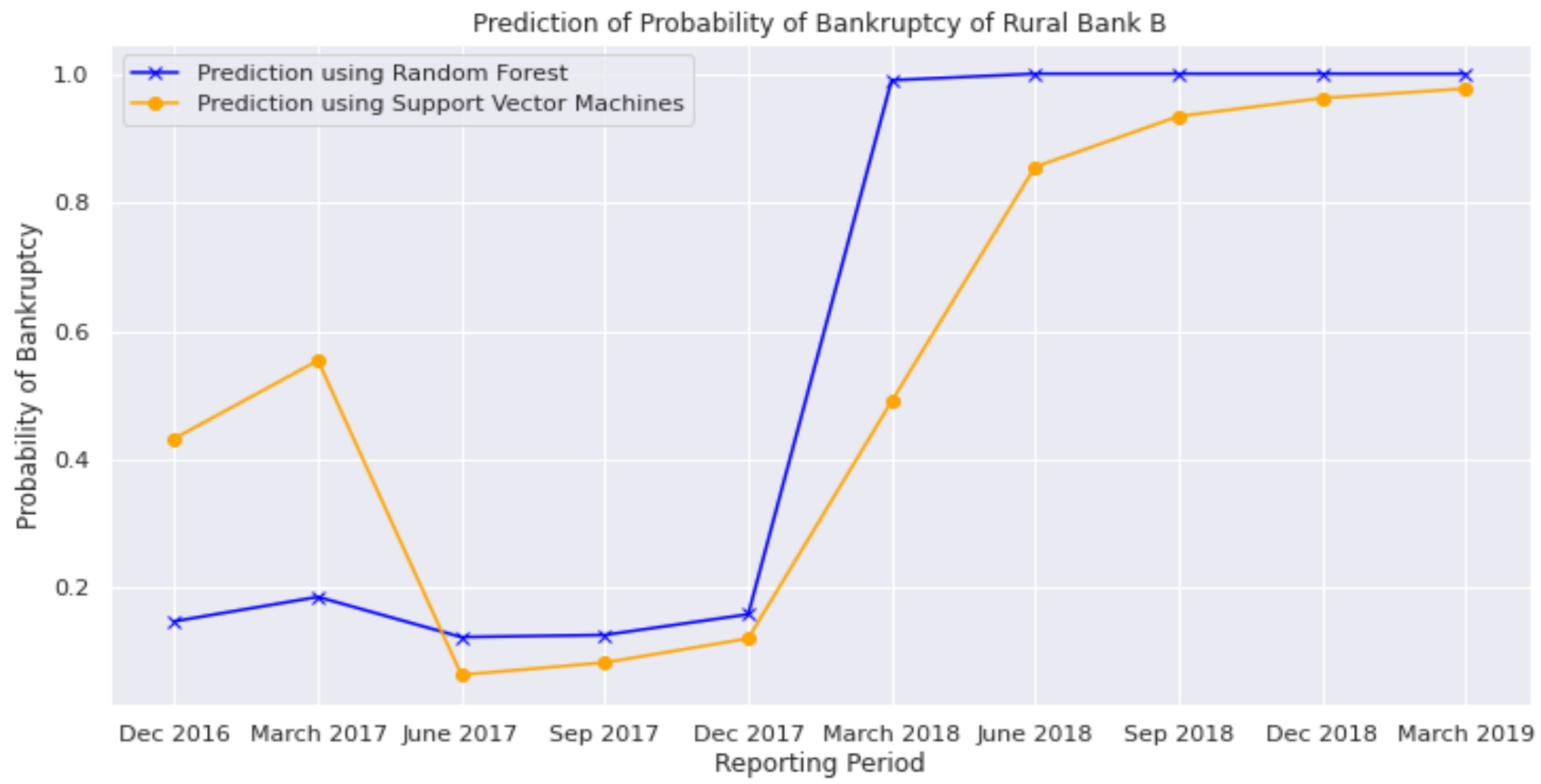

Figure 3. The Trend of Rural Bank B's Bankruptcy Probability Prediction.

As shown in Figure 3, rural bank B was predicted to fail in March 2017, but this prediction eased over the next three quarterly reports. Then, in March 2018, the chances of the bank going bankrupt increased significantly, and two months later, the bank was finally labelled "under intensive supervision."

The third sample, rural bank C, is located in Singkawan City, West Kalimantan. As of July 12, 2018, its business license was revoked. On April 5, 2018, the bank was placed "under special surveillance." Figure 4 illustrates the trend in predicting the probability of rural bank C's bankruptcy.

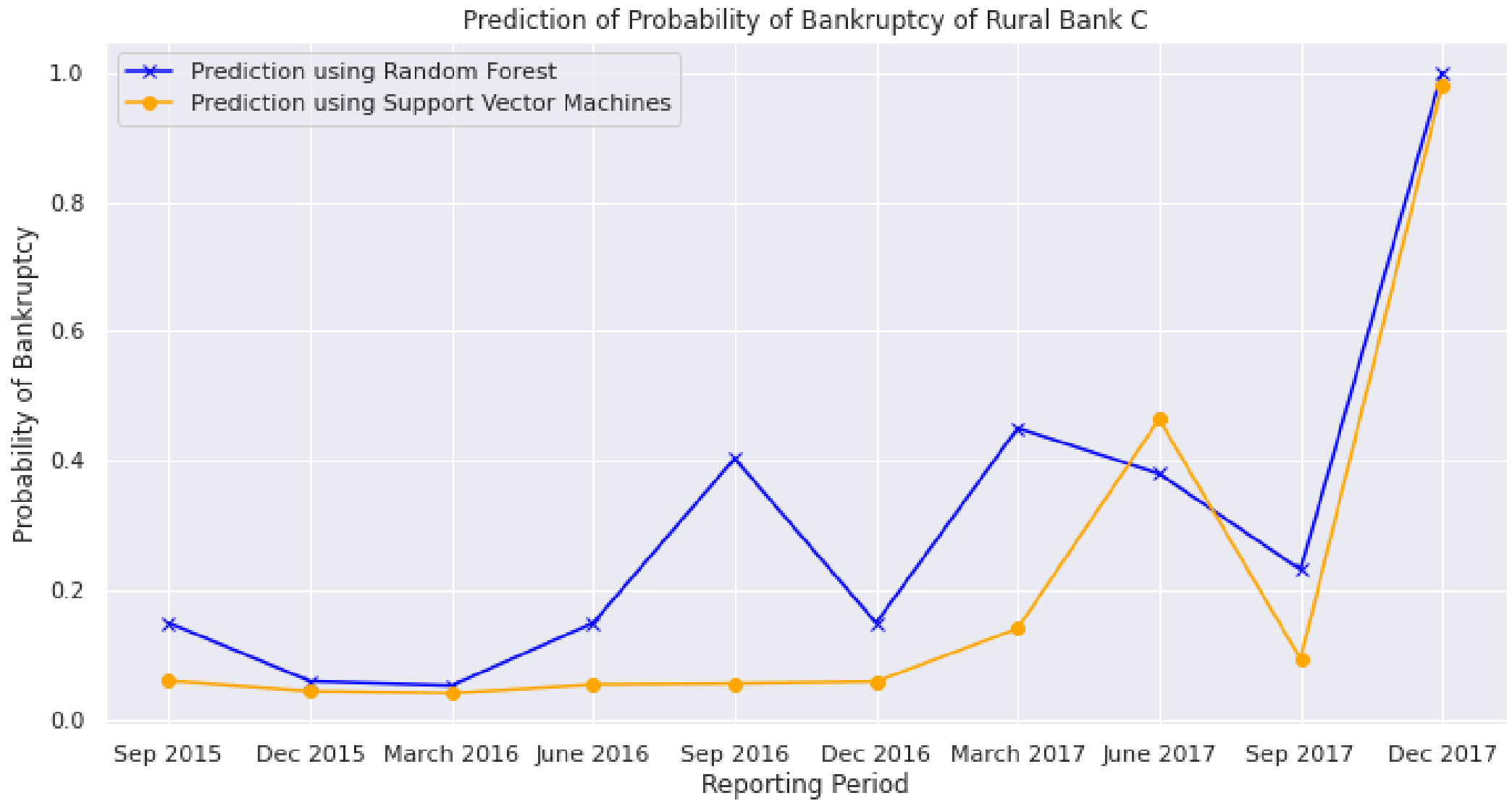

Figure 4. The Trend of Rural Bank C's Bankruptcy Probability Prediction.

Rural Bank C last filed financial reports in March 2018, despite its business license being revoked in July 2018. According to Figure 4, the trend probability of the bank failing remained stable until September 2017, before abruptly increasing at the end of the year. In September 2016 and March 2017, the predictions of the random forest model had a higher probability than those of the support vector machines model, although the predictions did not exceed 0.5.

The fourth and final sample is Rural Bank D, located in Solo, Central Java. On December 6, 2017, the bank's business license was revoked. Additionally, on May 10, 2017, the bank was placed "under special surveillance." Figure 5 illustrates the trend prediction probability of the rural bank D's bankruptcy.

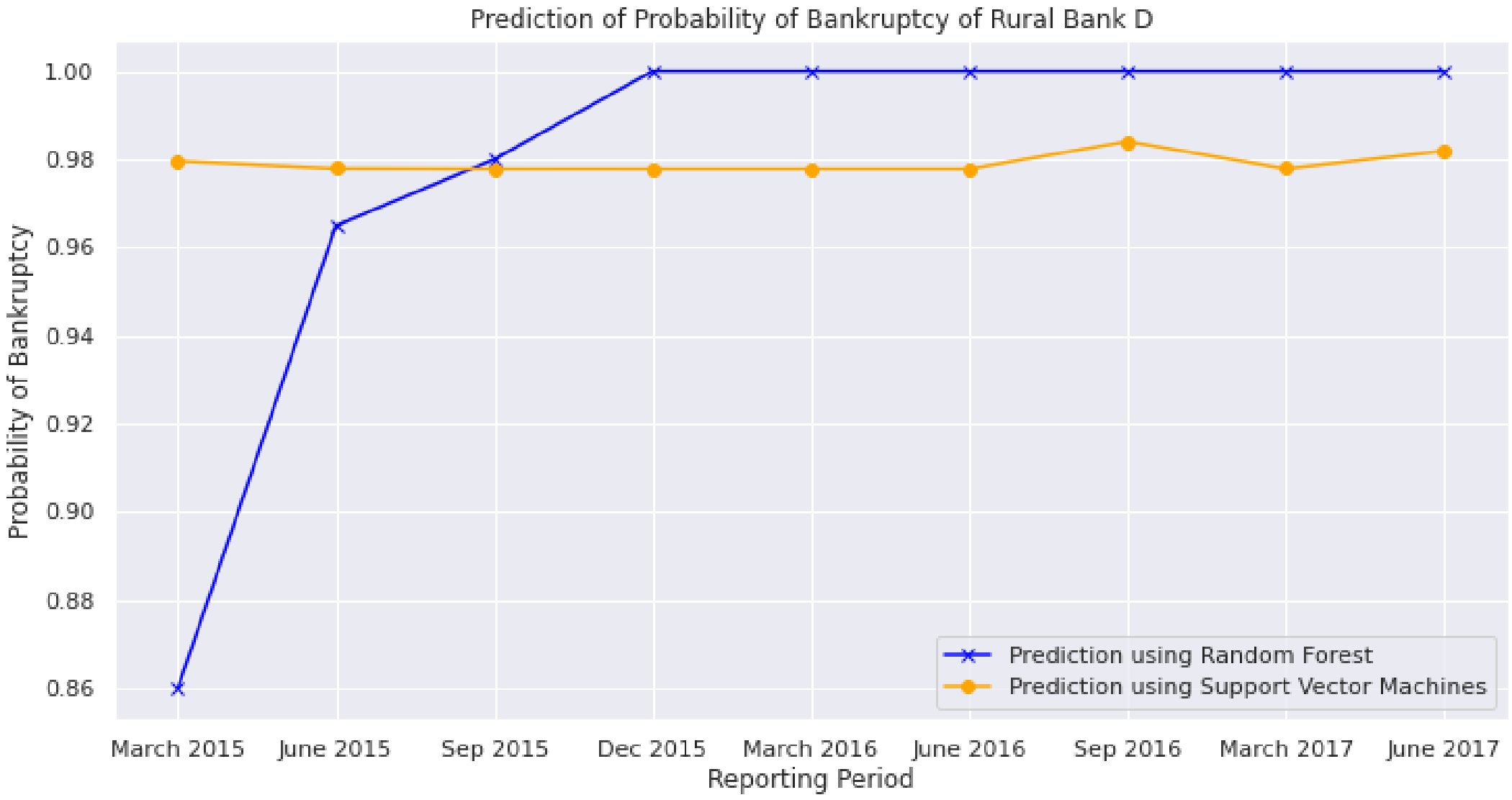

Figure 5. The trend of Rural Bank D's Bankruptcy Probability Prediction.

According to Figure 5, the support vector machine model accurately predicted Rural Bank D's bankruptcy in March 2015. This is indicated by the probability of bankruptcy which is predicted to be far greater than 0.5. The random forest model indicates that the bank will be bankrupt in June 2015.

Based on the trend in the probability of bank bankruptcy in four rural bank samples, it is believed that random forest and support vector machines work better when combined. On the other hand, both models can reasonably predict whether a rural bank will go bankrupt before it is labeled as "under intensive supervision" or "under special surveillance." Therefore, it is possible to use both models to determine the likelihood that a rural bank will go bankrupt, serving as a basis for the government to label a rural bank as "under intensive supervision" or "under special supervision."

## 7. Conclusion

Different data have dissimilar patterns and behaviors, so the same machine learning model will perform differently. This occurs in the performance of the machine learning model in commercial bank data and rural bank data. There are 20 predictor variables in commercial bank data, while rural bank data have only 5 predictor variables, even though both use the CAMEL ratio.

In commercial bank data, the random forest classifier achieves the highest accuracy of 90.91 per cent on the testing data, whereas the rural bank model achieves 100 per cent accuracy on the testing data using random forest, logistic regression, and support vector machine models.

Based on the historical trend analysis of the probability of default in rural bank data, it can be concluded that random forest and support vector machines are more effective when used in combination. This is based on the model's performance in predicting the probability of bankruptcy for the four rural bank data samples that have currently been liquidated or are in the process of liquidation. Based on the trend analysis of the four samples, the machine learning model can be used to obtain early warnings regarding the potential bankruptcy of the related rural banks.

## Acknowledgement and Disclaimer

SH receives financial support from the Indonesia Deposit Insurance Corporation (IDIC) through the 'LPS Call for Research 2020' grant scheme. ZR receives financial support from the University of Indonesia PUTI Q1 through NKB-488/UN2.RST/HKP.05.00/2022 grant scheme. We thank the reviewers from Indonesia Deposit Insurance Corporation (IDIC), i.e. Professor Iwan Jaya Aziz, Professor Irwan Adi Ekaputra, Dr. Piter Abdullah, Rimawan Pradiptyo, Ph.D., and Dr. Wahyoe Soedarmono, for their guidance, in-depth review, and fruitful comments and advice during the course of this research. We would also like to thank IDIC officials for their support. The results and findings of this research do not represent the view of IDIC. The IDIC did not provide any data during the process of this research. We declare that we have no competing interests with other parties.